\pdfoutput=1

\documentclass[11pt]{article}

\usepackage[final]{acl}

\usepackage{times}
\usepackage{latexsym}
\usepackage{amsmath,amssymb,amsfonts}
\usepackage{algorithmic}
\usepackage{graphicx}
\usepackage{textcomp}
\usepackage{xcolor}
\usepackage{algorithm}
\usepackage{booktabs}
\usepackage{makecell}
\usepackage{multirow}
\usepackage{subcaption}
\usepackage{tabularx} 
\usepackage[T1]{fontenc}

\usepackage[utf8]{inputenc}

\usepackage{microtype}

\usepackage{inconsolata}

\usepackage{graphicx}
\usepackage{courier}

\usepackage[most]{tcolorbox}    
\usepackage{lipsum}       

\usepackage{listings} 
\usepackage{xcolor} 
\usepackage{caption} 

\definecolor{codeblue}{rgb}{0.25,0.5,0.75}
\definecolor{codegreen}{rgb}{0.0,0.5,0.0}
\definecolor{codegray}{rgb}{0.5,0.5,0.5}
\definecolor{codeblue}{rgb}{0.25,0.5,0.75}
\definecolor{codegreen}{rgb}{0.0,0.5,0.0}
\definecolor{codegray}{rgb}{0.5,0.5,0.5}
\definecolor{lightgray}{rgb}{0.9,0.9,0.9}
\definecolor{highlight}{rgb}{0.8,0.95,0.8} 

\lstdefinestyle{mystyle}{
    backgroundcolor=\color{white},   
    commentstyle=\color{codegreen},
    keywordstyle=\color{codeblue},
    numberstyle=\tiny\color{codegray},
    stringstyle=\color{codeblue},
    basicstyle=\ttfamily\scriptsize,
    breakatwhitespace=false,         
    breaklines=true,                 
    captionpos=b,                    
    keepspaces=true,                 
    numbers=left,                    
    numbersep=5pt,                  
    showspaces=false,                
    showstringspaces=false,
    showtabs=false,                  
    tabsize=2
}

\title{LLM-Enhanced Self-Evolving Reinforcement Learning for Multi-Step E-Commerce Payment Fraud Risk Detection}

\author{
 \textbf{Bo Qu\textsuperscript{\ensuremath{\dagger}}\thanks{Corresponding Author.}},
 \textbf{Zhurong Wang\textsuperscript{\ensuremath{\dagger}}},
 \textbf{Daisuke Yagi\textsuperscript{\ensuremath{\mathsection}}},
 \textbf{Zhen Xu\textsuperscript{\ensuremath{\ddagger}}},
 \textbf{Yang Zhao\textsuperscript{\ensuremath{\dagger}}},
 \textbf{Yinan Shan\textsuperscript{\ensuremath{\dagger}}},
 \textbf{Frank Zahradnik\textsuperscript{\ensuremath{\dagger}}}
\\
 \textsuperscript{\ensuremath{\dagger}}eBay,
 \textsuperscript{\ensuremath{\ddagger}}University of Chicago,
 \textsuperscript{\ensuremath{\mathsection}}Etsy
\\
\texttt{{boqu@ebay.com}}
}

\begin{document}
\maketitle
\begin{abstract}
This paper presents a novel approach to e-commerce payment fraud detection by integrating reinforcement learning (RL) with Large Language Models (LLMs). By framing transaction risk as a multi-step Markov Decision Process (MDP), RL optimizes risk detection across multiple payment stages. Crafting effective reward functions, essential for RL model success, typically requires significant human expertise due to the complexity and variability in design. LLMs, with their advanced reasoning and coding capabilities, are well-suited to refine these functions, offering improvements over traditional methods. Our approach leverages LLMs to iteratively enhance reward functions, achieving better fraud detection accuracy and demonstrating zero-shot capability. Experiments with real-world data confirm the effectiveness, robustness, and resilience of our LLM-enhanced RL framework through long-term evaluations, underscoring the potential of LLMs in advancing industrial RL applications.
\end{abstract}

\section{Introduction}

\begin{figure*}[t]
\centering
\includegraphics[width=0.99\textwidth]{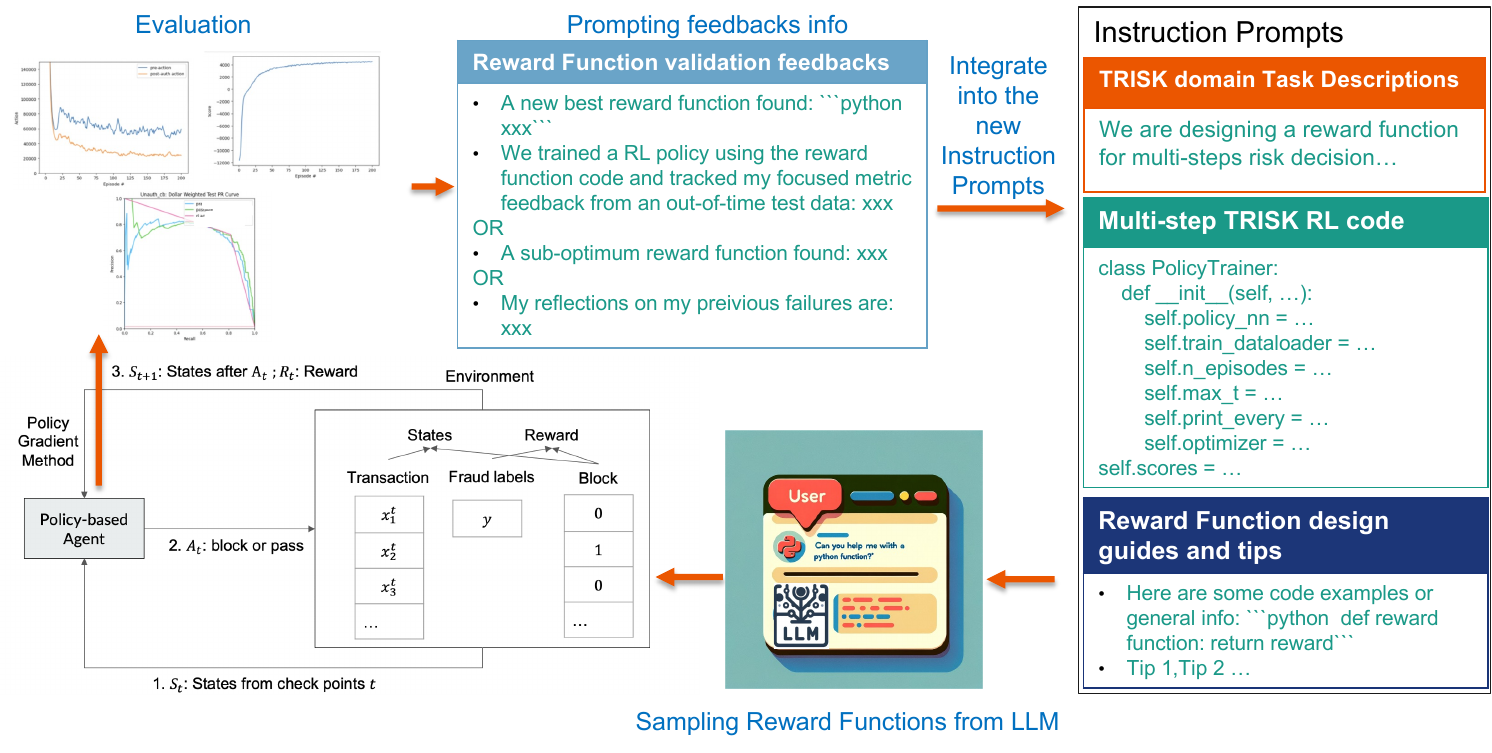}  
\caption{The LLM enhanced self-improving RL framework overview. It takes in the task description/instructions, the RL source code, and the example human-designed reward function as the context to generate an executable reward function. We designed an evolutionary algorithm to allow the LLM to evolve the reward function design based on feedback on the performance of the RL agent.}
\label{fig:LLM-RL-overview}
\end{figure*}

\begin{figure}[t]
\centering
\includegraphics[width=0.4\textwidth]{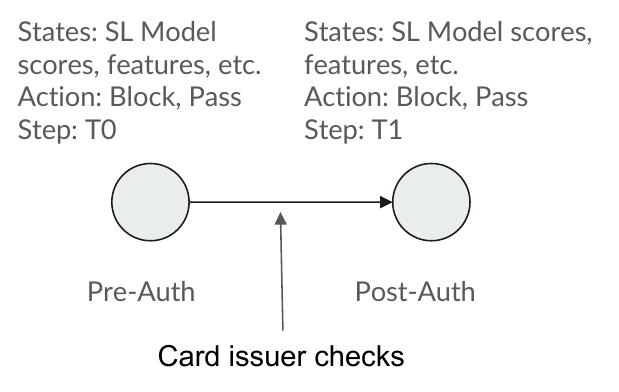}  
\caption{Imagine the buyer transaction risk decision checkpoints pipeline as a Markov Decision Process.}
\label{fig:MC}
\end{figure}

The advancement of LLMs has been remarkable, exemplified by notable developments such as the top-notch model API \cite{openai2023gpt} and state-of-the-art open-source models \cite{dubey2024llama3herdmodels} \cite{jiang2023mistral} \cite{jiang2024mixtral} \cite{team2024gemma} \cite{guo2024deepseek}. These breakthroughs have propelled LLMs to new heights in various tasks, reaching or even surpassing human capabilities in code generation \cite{chen2021evaluating}, logical reasoning \cite{kojima2022large}, and task planning \cite{shen2024hugginggpt}. The integration of these advanced capabilities into the domain of e-commerce payment fraud detection presents an exciting frontier for exploration.

Meanwhile, RL has shown its effectiveness in optimizing nondifferential goals and innovating decision strategies in response to environmental changes \cite{sutton2018reinforcement} \cite{russell2010artificial}. Its application in the financial fraud risk domain has seen various approaches, from modeling the sequence of transactions from a single credit card to considering each transaction as a discrete step in a MDP \cite{mead2018detecting} \cite{vimal2021application}. Other studies have explored the application of RL in fraud risk alerting systems \cite{shen2020deep} and discussed its potential without detailed propositions \cite{el2017fraud}. While supervised learning (SL) remains prevalent in static fraud detection, it struggles to model sequential dependencies between decision stages and directly optimize business metrics like precision-recall tradeoffs – limitations that RL naturally addresses through reward-driven optimization.

The confluence of LLM’s semantic capabilities with RL has sparked interest, particularly in using LLMs as a reward shaper for RL. This innovative approach includes directly feeding the context of the environment to LLMs for action and reward processing \cite{kwon2023reward}, using LLMs to define the parameters of the reward function \cite{yu2023language}, or even to design whole rewards function codes \cite{ma2023eureka}. These efforts have mainly focused on gaming agents and robotic task control, inspiring our exploration into e-commerce payment fraud detection.

E-Commerce payment fraud presents a dynamic challenge necessitating advanced decision-making across three key stages: 1) \textit{Pre-authorization} (Pre-auth) where our platform screens transactions before card issuers' risk assessment, 2) \textit{Issuer check} where card networks validate payment credentials, and 3) \textit{Post-authorization} (Post-auth) where we conduct final risk evaluation after issuer approval. Traditional SL approaches operate isolated classifiers at each stage, failing to model the sequential interdependencies and business constraints (e.g., needing to block more potential frauds during Pre-auth to avoid issuer penalties). This fragmentation leads to suboptimal precision-recall balance and excessive manual reviews. RL’s strength in constrained sequential optimization makes it uniquely suited to maximize cumulative fraud prevention while respecting stage-specific requirements.

In response, we propose a cutting-edge RL framework that harnesses the power of LLM to autonomously evolve and refine decision-making processes in the payment risk domain, a first in this field. Our contributions are summarized as follows:

\textbf{LLM-based Reward Function Generation for RL}: We introduce a framework using LLMs to autonomously create reward functions that directly optimize precision-recall metrics in the payment risk domain, outperforming human-designed rewards. It uses an evolutionary algorithm for iterative refinement based on RL agent feedback, supporting few-shot/zero-shot creation with/without prior examples. The general process is shown in Figure~\ref{fig:LLM-RL-overview}.

\textbf{Transaction Risk Detection as Constrained MDP}: We redefine transaction risk detection as a multistep MDP with stage-specific constraints, solved using policy-based RL like REINFORCE. By integrating transaction stages into a coherent framework (see Figure~\ref{fig:MC}) and aggregating reward signals across stages (detailed in Figure~\ref{fig:RL-fw}), our method outperforms SL’s surrogate loss functions through direct optimization of business objectives.

Our research, supported by extensive experiments with real-world e-Commerce transaction data, demonstrates significant improvements in fraud detection performance compared to the existing SL models on our payment system.

\section{Methodology}

\subsection{Designing the MDP and RL Framework}
We model the e-commerce transaction process as a finite-horizon MDP, visualized in Figure~\ref{fig:MC}. The system generates state signals from both legacy SL risk model scores and transaction stage indicators (Pre-auth, Post-auth). While there are also many transactional features that can be used as state signals, our experiments primarily use SL scores for state representation due to their proven predictive value leveraging all the features, the framework can theoretically incorporate any transactional features available at each stage. The policy agent uses these state signals to decide between risk responses ("block" or "allow"), with the MDP structure enabling sequential decision-making that supervised learning cannot naturally accommodate.

\begin{figure}[t]
    \centering
    \begin{minipage}{0.48\textwidth}
        \centering
        \includegraphics[width=\linewidth]{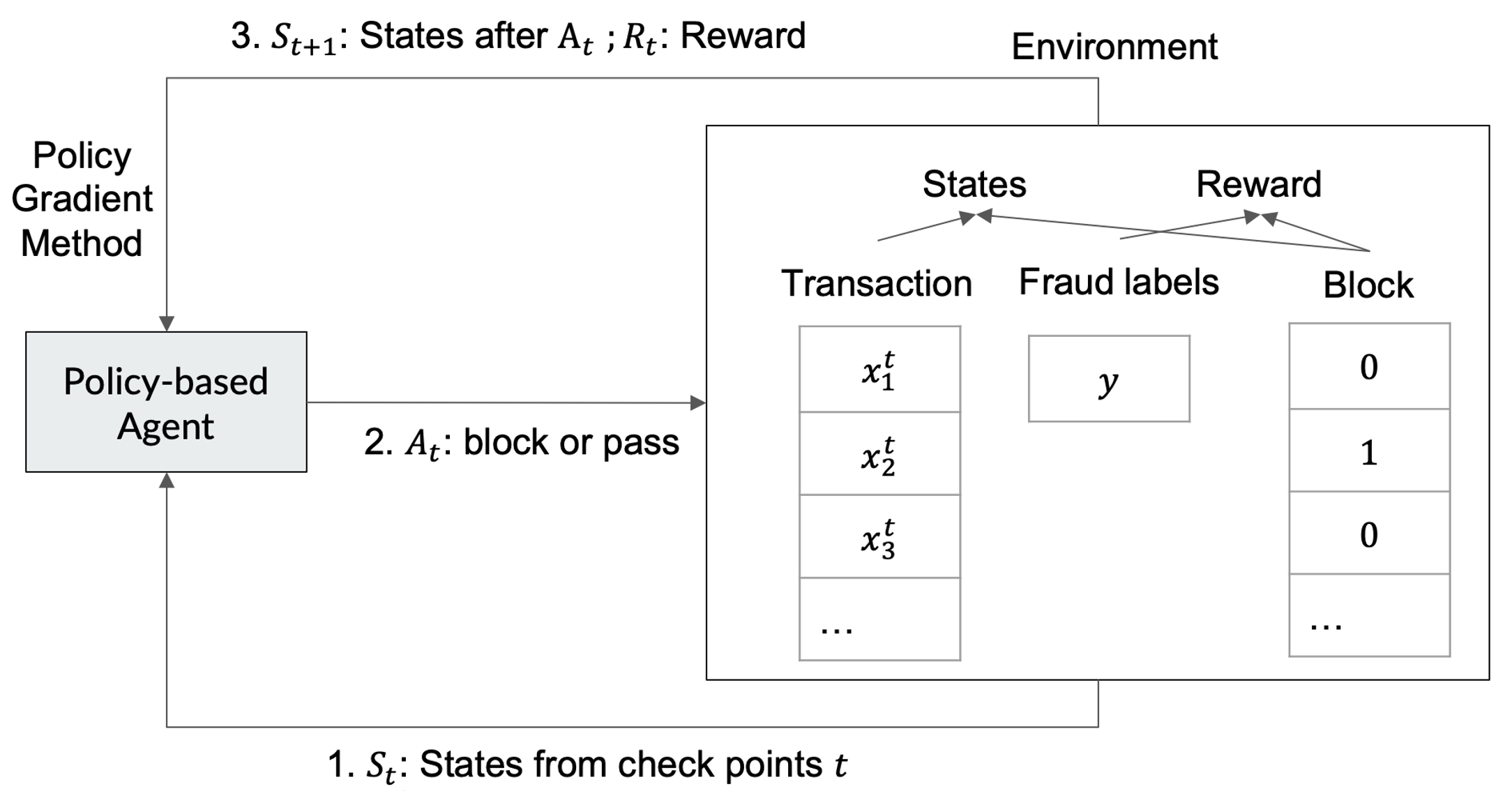}
        \caption{TRISK MDP framework with staged decision points. States incorporate SL risk scores and stage indicators.}
        \label{fig:RL-fw}
    \end{minipage}\hfill
\end{figure}

The agent-environment interaction (Figure~\ref{fig:RL-fw}) defines:
\begin{itemize}
    \item $\mathcal{S}_i =$ SL scores \textit{and stage indicators} at step $i$
    \item $\mathcal{A}_i =$ possible risk responses (block, pass)
    \item $\mathcal{R}_i = R(\mathcal{S}_i,\mathcal{A}_i)$, the reward function
\end{itemize}
We maximize the business-driven objective:
\begin{align}
\label{equ:opt_target}
\text{Maximize } & \; \text{\$TP} - \text{\$FP} \\
\text{subject to } & \; \text{\$TP}_{\text{stage 1}} > \text{\$TP}_{\text{stage 2}} \nonumber
\end{align}

where dollar-wise \$TP-\$FP optimization directly meets the theoretical goal of our risk business, which corresponds to maximizing fraud prevention while minimizing Loss of the Gross Merchandise Value (GMV) from false positives. The decreasing \$TP constraint reflects practical fraud patterns where early detection captures higher-value fraud attempts.

We employ offline RL with policy gradient methods (REINFORCE \cite{williams1992simple}) using historical transaction data. To address offline evaluation challenges, we firstly try to train with enough amount of transaction data, and secondly we validate policies on extended test periods (6+ months) demonstrating consistent performance before production deployment.

\subsection{Human Reward Function Design}

While Equation~\ref{equ:opt_target} captures core business objectives, real-world operations require balancing specific precision-recall trade-offs across transaction categories. Here we figured out the reward design that achieve this implicitly through directly considering the optimization constraints instead of the optimization goal itself. By transforming operational constraints into differentiable objectives through algebraic manipulation, we found that it naturally merges into the optimization goal considering the precision block level.

\paragraph{Precision Constraint based Reward Function}

Business requirements ($\text{\$TP}_{\text{stage 1}} >\text{\$TP}_{\text{stage 2}}$) dictate precision thresholds $\alpha_i$ per stage, with $\alpha_1 < \alpha_2$ enforcing stricter precision in later stages. Hence we assume the blocking precision inequality in stage $i$:
\begin{equation}
\frac{\text{\$TP}_{i}}{\text{\$TP}_{i} + \text{\$FP}_{i}} > \alpha_i
\end{equation}
we derive the reward function through Lagrangian relaxation:
\begin{equation}
\label{equ:rw_1}
R_{\text{precision}}^{i}(s,a)= (1-\alpha_i)\text{\$TP}_i - \alpha_i\text{\$FP}_i > 0
\end{equation}
Maximizing this implicitly maximizing (\$TP - \$FP) while maintaining stage-wise constraints by introducing the coefficients in front these terms, derived naturally from the inequality above.

While effective, these human-designed rewards require careful parameter tuning, and in theory there could be more effective designs that need more human efforts to explore. Therefore, we proposed a LLM-enhanced framework automates this exploration by incorporating the specifications of policy performance feedback in natural language, to further enhance the RL reward signals.

\subsection{LLM-based Reward Function Optimization}

We propose a framework using LLMs to dynamically optimize reward functions in our evolving RL algorithm for e-commerce payment fraud detection.

\subsubsection{Algorithm Overview}

Our method, detailed in Algorithm~\ref{alg:LLM-RL alg}, employs Enhanced LLM-based Reward Optimization for RL agents, evolving the reward function to boost decision-making. The cycle includes:

1. Initialization with environment $\mathcal{E}$, baseline model $\mathcal{M}_b$, and metrics.

2. Generation of reward candidates by an LLM, guided by temperature for novelty.

3. Validation and use of candidates to train RL agents for fraud detection.

4. Evaluation of detection accuracy and impact, informing reward success.

5. Self-Reflection: top functions update the LLM context; failures refine iterations.

6. Repeat steps 2-5 until iteration or convergence.

To ensure the executability of generated reward functions, we implement a two-step validation process: (1) incorporating basic reward function structure requirements in the prompts, and (2) using preliminary code checks to confirm that generated functions fit the required structure. If a function fails these checks, the LLM regenerates it during the sampling phase, significantly reducing unexecutable cases, without human in the loop.

\subsubsection{Customized In-Context Prompt}

The initial and iterative instructions provided to the LLM are critical to the success of our algorithm. We construct a domain-specific prompt that outlines the objectives of the reward function, incorporates the RL environment framework, and includes basic requirements and examples. As shown in Figure \ref{fig:LLM-RL-overview}, the prompt is dynamically updated with feedback loop information, allowing the LLM to adapt its generative process to the evolving requirements of the fraud detection task. Examples of prompts are shown in the following boxes, with more detailed content in Appendix \ref{appendix:Prompts Design}.

\begin{tcolorbox}[title=Initial Instruction Prompts, colframe=gray, colback=white, breakable, fonttitle=\large\bfseries, fontupper=\small]
You are a reward engineer trying to write reward functions to solve reinforcement learning tasks as effectively as possible. Your goal is to:
(1) ...
(2) ...
The goal of my task is: ..., my codes framework of input data as states and train my policy is shown in the code: ```python \{...\} ```.

Your reward function should use useful variables from my codes framework as inputs. As some examples,
here are some example reward functions proposed by humans: ```python \{...\} ```,
and here is the best reward function signature so far: ```python \{...\} ```
...
The output of the reward function should consist of:
(1) ...
(2) ...
...
\end{tcolorbox}

\begin{tcolorbox}[title=Feedback Prompts, colframe=gray, colback=white, breakable,fonttitle=\large\bfseries, fontupper=\small]
We trained a RL...:

1. RL Agent Training info: ...

2. Test evaluation info: ...

Moreover, the ratio between the bad GMV blocked by first step and the bad GMV blocked by second step is: \{...\}/\{...\} ...

Error occurred during training: \{...\}

Error occurred during evaluating: \{...\}
\end{tcolorbox}

\subsubsection{Zero-shot and Few-shots setups}

Our approach supports both zero-shot and few-shot capabilities. In the zero-shot setup, the algorithm generates reward functions based on general component descriptions rather than predefined human-designed functions. For the few-shot setup, detailed examples of human-crafted reward functions are included in the prompt, allowing the model to reference specific code and build on these exemplars.

Feedback and success metrics play a crucial role in optimizing the reward function, especially in zero-shot scenarios. Feedback comprises policy evaluation results, such as precision-recall on test data, error reports, and comparative evaluations of previous best and sub-optimal rewards. Importantly, in cases where no sub-optimal reward is found, a reflection process allows the LLM to summarize insights from failed reward functions, integrating this experience into instructions for subsequent iterations, as described in line 26 of Algorithm~\ref{alg:LLM-RL alg}. This reflective feedback is vital for zero-shot cases.

\begin{algorithm}
\caption{LLM-based Reward Function Optimization for RL Agent}
\label{alg:LLM-RL alg}
{\tiny
\begin{algorithmic}[1]
\REQUIRE $N_{iter}, N_{samples}, N_{episodes}, \theta_{recall}, R_{scores}$
\STATE Initialize environment $\mathcal{E}$, baseline model $\mathcal{M}_b$, and evaluation parameters
\STATE $f_{best} \leftarrow$ InitializeBestRewardFunction(), Initialize LLM temperature parameters
\STATE Load baseline model performance and set evaluation criteria
\FOR{$iter = 1$ to $N_{iter}$}
    \STATE Initialize feedback and success lists: $feedbacks, success$
    \STATE Update LLM temperature based on feedback loop criteria
    \FOR{$sample\_i = 1$ to $N_{samples}$}
        \STATE Sample and validate $f_{reward}^{sample\_i}$ using LLM with temperature control
        \IF{valid $f_{reward}^{sample\_i}$}
            \STATE Save $f_{reward}^{sample\_i}$, proceed to training
        \ELSE
            \STATE Re-sample $f_{reward}^{sample\_i}$
        \ENDIF
    \ENDFOR
    \FOR{each valid $f_{reward}^{sample\_i}$}
        \STATE $\mathcal{A}_i \leftarrow$ TrainAgent$(\mathcal{E}, f_{reward}^{sample\_i}, N_{episodes})$
        \STATE $feedback_i, success_i \leftarrow$ EvaluateAgent$(\mathcal{A}_i, \mathcal{M}_b, \theta_{recall}, R_{scores})$
        \STATE Append $feedback_i$ to $feedbacks$ and $success_i$ to $success$
    \ENDFOR
    \STATE Update $f_{best}$ based on evaluation results, Update LLM temperature and instructions for next iteration based on feedback loop outcomes
    \IF{new $f_{best}$ found}
        \STATE Update system instructions for LLM to include new best reward function details
    \ELSIF{sub-optimal reward function found}
        \STATE Update system instructions for LLM to include sub-optimal reward function details as feedback
    \ELSE
        \STATE Let LLM summarize reflections based on the failed reward functions info and include its experience into the instructions for next iteration
    \ENDIF
\ENDFOR
\end{algorithmic}
}
\end{algorithm}

\subsubsection{Interpretability of LLM-Generated Reward Functions}

While the proposed framework leverages LLMs to automatically evolve reward functions for RL agents, it is important to acknowledge that such LLM-generated reward functions inherently carry a degree of "black-box" behavior, especially in zero-shot settings. To enhance interpretability, we embed domain-specific contextual information into the prompts provided to the LLM.

In both zero-shot and few-shot reward function design prompts, we explicitly define domain-specific contexts such as key business metrics — \$TP, \$FP, \$TN, and \$FN — along with their implications in fraud detection (lines 6–9 in the prompt example below). These definitions are paired with optimization objectives and constraints within the domain context (lines 10–11), further reinforced by additional descriptions in the instruction prompts and feedback mechanisms detailed in Section 2.3.2. This structured context guides the LLM to generate reward functions that align closely with real-world business requirements. Take the zero-shot reward design as an example: in Listing~\ref{lst:zero_shot_rw}, the LLM incorporates terms such as \$FP and \$FN, indicating its understanding of the trade-offs between \$TP vs. \$FP and \$TN vs. \$FN. It also assigns higher weights to early-stage rewards (e.g., reward *= 1.2 at current\_step == 0 and reward *= 0.9 at current\_step == 1), reflecting the business requirement that detecting fraud earlier yields greater value.

\begin{tcolorbox}[title=Domain-Specific Context Prompts for Reward Function Design, colframe=gray, colback=white, breakable,fonttitle=\large\bfseries, fontupper=\small]

1. element in action either equals 0 or 1;

2. action == 1 means the transactions that were taken blocking action, action == 0 means the transactions that were taken pass action;

3. element in target either equals 0 or 1;

4. target == 1 means the transactions that are tagged as fraud risk, target == 0 means the transactions are not tagged as risk;

5. wgt is the tensor of dollarwise weight for each transaction;

6. e.g. ((action==1) \& (target==1) * wgt) means the tensor that have the True Postive GMV value where (action==1) \& (target==1); 

7. e.g. ((action==1) \& (target==0) * wgt) means the tensor that have the False Positive GMV value where (action==1) \& (target==0);

8. e.g. ((action==0) \& (target==0) * wgt) means the tensor that have the True Negative GMV value where (action==0) \& (target==0);

9. e.g. ((action==0) \& (target==1) * wgt) means the tensor that have the False Negative GMV value where (action==0) \& (target==1);

10. the general goal of this reward function is to drive the agent to increase True Postive GMV and True Negative GMV, decrease False Positive GMV and False Negative GMV;

11. this reward function need to drive the agent to block more potential True Postive GMV at the current\_step == 0 than at the current\_step == 1.
\end{tcolorbox}

Despite these efforts, certain aspects — such as why specific parameter choices lead to particular precision-recall outcomes on certain test data— remain difficult to fully interpret. Therefore, we complement the validation of the reward function design with long-term evaluations (Test L in Table~\ref{tab:DatasetInfo}), demonstrating the stability and practical effectiveness of the evolved reward functions over time.

\begin{figure}[ht!]
    \centering

    \begin{subfigure}{0.48\textwidth}
        \centering
        \begin{lstlisting}[language=Python, backgroundcolor=\color{highlight}, caption={Original zero-shot reward function design by Mixtral8X7B. The calculation of rewards and penalties in both steps is uniquely different compared to Equation~\ref{equ:rw_1} above.}, label={lst:zero_shot_rw}]
def get_reward(current_step, action, target, wgt):
    reward = (action * target * wgt).float()  
    if current_step == 0:
        reward *= 1.2
    elif current_step == 1:
        reward *= 0.9
    fn = ((1 - action) * target * wgt).float()
    reward -= fn * 0.5 
    fp = ((action * (1 - target) * wgt).float())
    reward -= fp * 0.1
    low_weight_penalty = (action * (wgt < 50)).float()
    reward -= low_weight_penalty * 0.005
    reward /= wgt
    return reward
        \end{lstlisting}
    \end{subfigure}

    \begin{subfigure}{0.48\textwidth}
        \centering
        \begin{lstlisting}[language=Python, backgroundcolor=\color{highlight}, caption={Original few-shot reward function design by Mixtral8X7B. This design introduces unique reward terms compared to Equation~\ref{equ:rw_1} above, rather than simply adjusting the parameters of the human-designed version.}, label={lst:few_shot_rw}]
def get_reward(current_step, action, target, wgt):
    gamma_positive = 1.15
    gamma_negative = 0.9
    alpha = 1.2
    reward = 0
    if current_step == 0:
        reward = gamma_positive * (
            ((action == 1) & (target == 1)) * wgt - 
            ((action == 1) & (target == 0)) * (alpha * 0.005) * wgt - 
            0.15 * ((action == 0) & (target == 1)) * wgt  
        )
    elif current_step == 1:
        reward = gamma_negative * (
            ((action == 1) & (target == 1)) * wgt -  
            ((action == 1) & (target == 0)) * (alpha * 0.002) * wgt -  
            0.10 * ((action == 0) & (target == 1)) * wgt 
        )
    return reward
        \end{lstlisting}
    \end{subfigure}
    \caption{Reward function designs evolved by Mixtral8X7B in different contexts: Listing (1) Zero-shot context, Listing (2) Few-shot context.}
    \label{fig:reward_comparison}
\end{figure}

\subsubsection{Generalizability Discussion}
State-of-the-art approaches, such as those presented by \cite{ma2023eureka}, have employed evolutionary loops to demonstrate the robustness of these methods in optimizing RL training processes within different robotics tasks. However, these frameworks are primarily tailored to the specific data and scenarios encountered in robotics, limiting their direct applicability to our domain. Therefore, our work introduces this novel adaptation of evolutionary loops for tasks in e-commerce risk detection, for the first time. By doing so, we first demonstrate that this evolutionary reward design loop, leveraging LLMs, can be effectively generalized to e-commerce payment fraud scenarios. Theoretically, this approach can also be extended to other RL tasks within this domain that share similar data structures and objectives.

\section{Experiments}


\subsection{Datasets and Evaluation Metrics} 

We used real-world transaction data focusing on Pre-auth and Post-auth stages. SL models (gradient boost machines) scores $\mathcal{S}_i = \{Scr_{i0},\cdots,Scr_{ij}\}$ on the 2 stages, and stage indicators, represented the RL state. Data were split, labeled with our key fraud signals, and evaluated on out-of-time test sets. Table \ref{tab:DatasetInfo} shows dataset details. Test S, with 522K transactions, allows for quick performance comparisons but may introduce more variance due to its size. In contrast, Test L, with 6.17M transactions, offers more robust validation.

\begin{table}[t]
\small
\caption{Experiment Datasets.}
\centering
\begin{tabular}{lp{1.8cm}p{1.5cm}p{1.5cm}} 
    \toprule
    Dataset & Time Window & Total & Fraud Label \\
    \midrule
    Train & 2023-09-01 to 2023-09-14 & 2,136,590 & 28,226 \\ 
    Test S & 2023-09-15 to 2023-09-30 & 522,105 & 825 \\
    Test L & 2023-11-01 to 2024-04-30 & 6,174,069 & 7,834 \\
    \bottomrule
\end{tabular}
\label{tab:DatasetInfo}
\end{table}

\begin{table}[t]
\small
\caption{Performance of Policy Agent vs. Baseline, on Test S.}
\centering
\begin{tabular}{lccc}
    \toprule
    \makecell[l]{Recall \\ Levels} & \makecell[c]{Baseline \\ \$Prec} & \makecell[c]{RL Agent \\ \$Prec} & \makecell[c]{Bad GMV \\ Catch Ratio} \\
    \midrule
    @80\%   & 66.57\% & \textbf{69.65\%} & \textbf{9.79} \\ 
    @85\%   & 58.79\% & \textbf{64.22\%} & \textbf{15.32} \\ 
    @90\%   & 51.27\% & \textbf{55.7\%}  & \textbf{13.36} \\ 
    \bottomrule
\end{tabular}
\label{tab:Part1readout}
\end{table}

We assess performance using a metric for dollar-wise precision (\$Precision) at key dollar-wise recall (\$Recall) levels, calculated by our main fraud label. This metric is crucial as it aims to maximize legitimate GMV by minimizing \$FP transaction values at a given risk level. For the RL agent scores, we find combinations of blocking score thresholds across two stages to achieve the desired overall \$Recall, then observe the \$Precision. For the baseline model, we use the Pre-auth SL model score, which is most commonly employed by the policy, to observe this metric. Due to the complexity of human analysis in business practice, no cross-stage policy has been designed previously using SL model scores as a baseline. Which is also why we need to propose our RL solution in the first place. 

\subsection{Experimental Results and Analysis}

\begin{table*}[t]
\centering
\small
\caption{Zero-shot and Few-shot Performance Comparison of LLMs in LLM+RL Approach, on Test S.}
\label{tab:performance_comparison}
\begin{tabular}{lccp{1.5cm}p{1.5cm}p{1.5cm}p{1.5cm}p{1.5cm}p{1.5cm}}
    \toprule
    \multirow{2}{*}{Recall Levels} & \multirow{2}{*}{Baseline \$Prec} & \multicolumn{3}{c}{Zero-shot Evolved RL agent \$Prec} & \multicolumn{3}{c}{Few-shot Evolved RL agent \$Prec} \\
    \cmidrule(lr){3-5} \cmidrule(lr){6-8}
    & & Mixtral-8x7B & Gemma7B & LLaMa-3-8B & Mixtral-8x7B & Gemma7B & LLaMa-3-8B \\
    \midrule
    @80\%  & 66.57\% & \textbf{72.71\%} & \textbf{73.27\%} & \textbf{72.86\%} & \textbf{73.41\%} & \textbf{73.53\%} & \textbf{73.74\%} \\ 
    @85\%  & 58.79\% & \textbf{69.62\%} & \textbf{65.42\%} & \textbf{69.40\%} & \textbf{70.73\%} & \textbf{69.87\%} & \textbf{71.70\%} \\ 
    @90\%  & 51.27\% & \textbf{57.42\%} & \textbf{53.65\%} & \textbf{57.06\%} & \textbf{58.00\%} & \textbf{56.93\%} & \textbf{55.90\%} \\ 
    \bottomrule
\end{tabular}
\end{table*}

\textbf{Part 1: Human-designed Reward Function}:
In the first segment, a single RL agent was trained using a 3-layer neural network with dimensions [8, 32, 8], incorporating dropout layers and GELU activation functions. The model processed a four-dimensional input consisting of representative scores from legacy SL models, which served as the state representation. The output was the probability of taking the "block" action. Training was conducted using the REINFORCE algorithm with the Adam optimizer.

Multiple trials stabilized results, Table \ref{tab:Part1readout} shows enhanced performance and risk detection efficiency, with the agent blocking more fraudulent GMV in the Pre-auth stage. 

All training in part 1 was performed on a machine equipped with a single V100 GPU (32GB VRAM), 32 CPU cores, and 450GB of RAM. With our current implementation, iterating over 200 training epochs — generally sufficient for observing convergence in our experiments — took approximately 20 minutes per epoch. Each iteration involved processing the full training dataset, as detailed in Table~\ref{tab:DatasetInfo}.

\textbf{Part 2: LLM-enhanced Reward Function}:
We employed LLM-enhanced rewards using models like Mixtral-8x7B, LLaMa-3-8B, and Gemma7B. Experiments included zero-shot and few-shot setups with varying LLM prompts. Algorithm~\ref{alg:LLM-RL alg} parameters included $N_{iter} \approx 60$, $N_{samples} \approx 10$, $N_{episodes}\approx 150$, and $\theta_{recall}\in [80\%,85\%,90\%]$. Results are in Table \ref{tab:performance_comparison}.

Zero-shot scenarios used descriptive prompts without reward function examples, leading to competitive reward designs, as shown in Listing~(\ref{lst:zero_shot_rw}). Few-shot scenarios also allowed LLMs to modify and create reward functions, as shown in Listing~(\ref{lst:few_shot_rw}), improving performance metrics. Zero-shot setups required more iterations, indicating optimization potential, but overall, LLM-enhanced approaches showed adaptability and innovation.

Each complete training iteration, encompassing LLM inference, RL agent training, and performance evaluation, required approximately 40 minutes. All experiments in part 2 were conducted on a machine equipped with 2 V100 GPUs (32GB VRAM), 32 CPU cores, and 450GB of RAM, with LLMs loaded in 4-bit precision ($load\_in\_4bit=True$) to reduce VRAM consumption. The primary computational bottlenecks were identified as LLM inference and policy evaluation. These components represent key areas for future optimization in the implementation pipeline.

\textbf{Part 3: Long-term Evaluation}:
To test RL agent robustness over time, we extended evaluation on Test L covering six additional months. Using the same RL agent, we analyzed performance with \$Prec metric against a baseline model at similar \$Recall thresholds for all LLMs in both zero-shot and few-shot scenarios.

\begin{figure}[t]
    \centering
    \begin{minipage}{0.5\textwidth}  
        \centering
        \includegraphics[width=\linewidth]{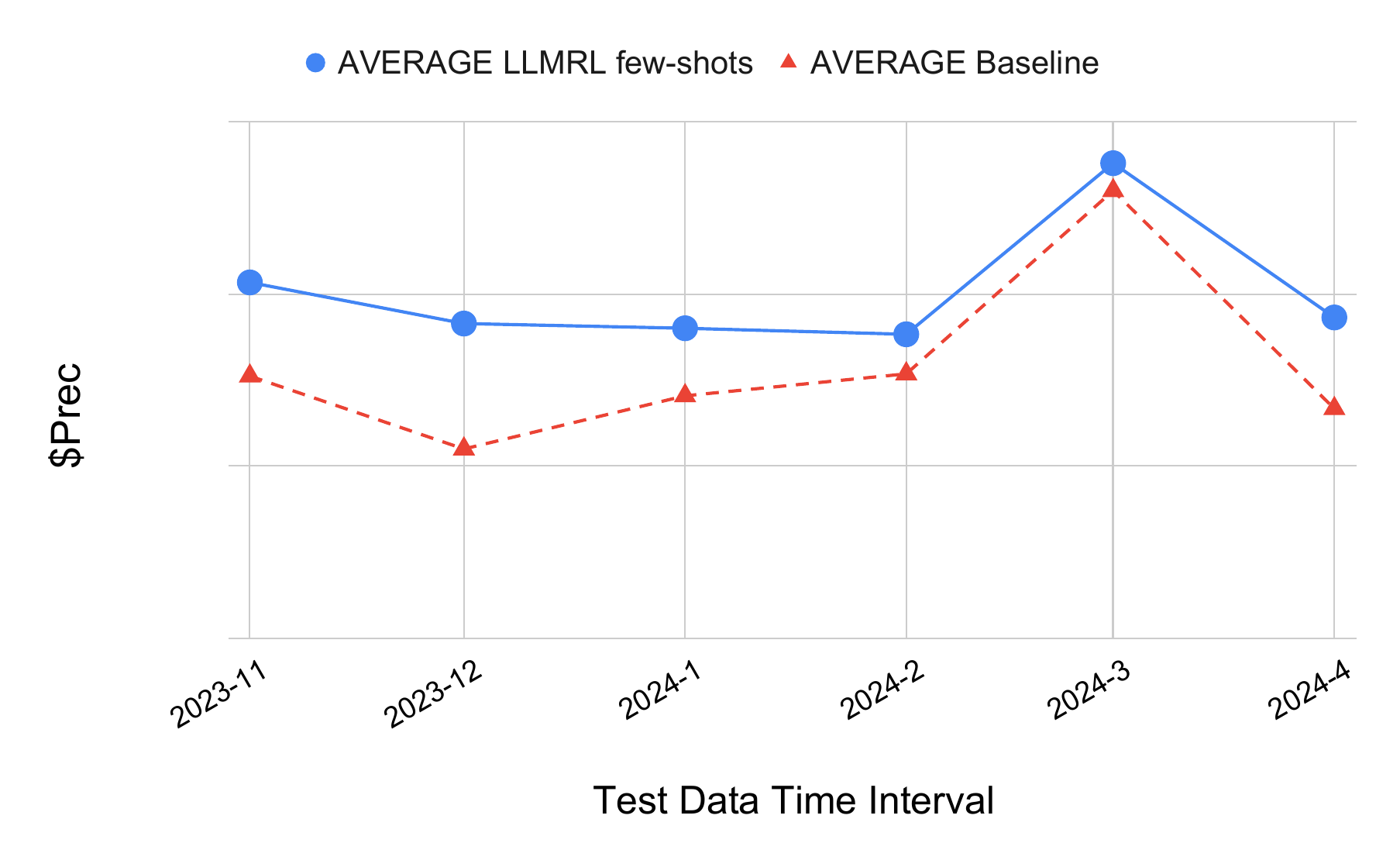}  
        \caption{Averaged blocking \$Prec@\$Recall from 3 LLM guided RL agents, in the few-shots scenario, on Test L.}
        \label{fig:LLMRL_LT_fs}
    \end{minipage}\hfill
    \begin{minipage}{0.5\textwidth}  
        \centering
        \includegraphics[width=\linewidth]{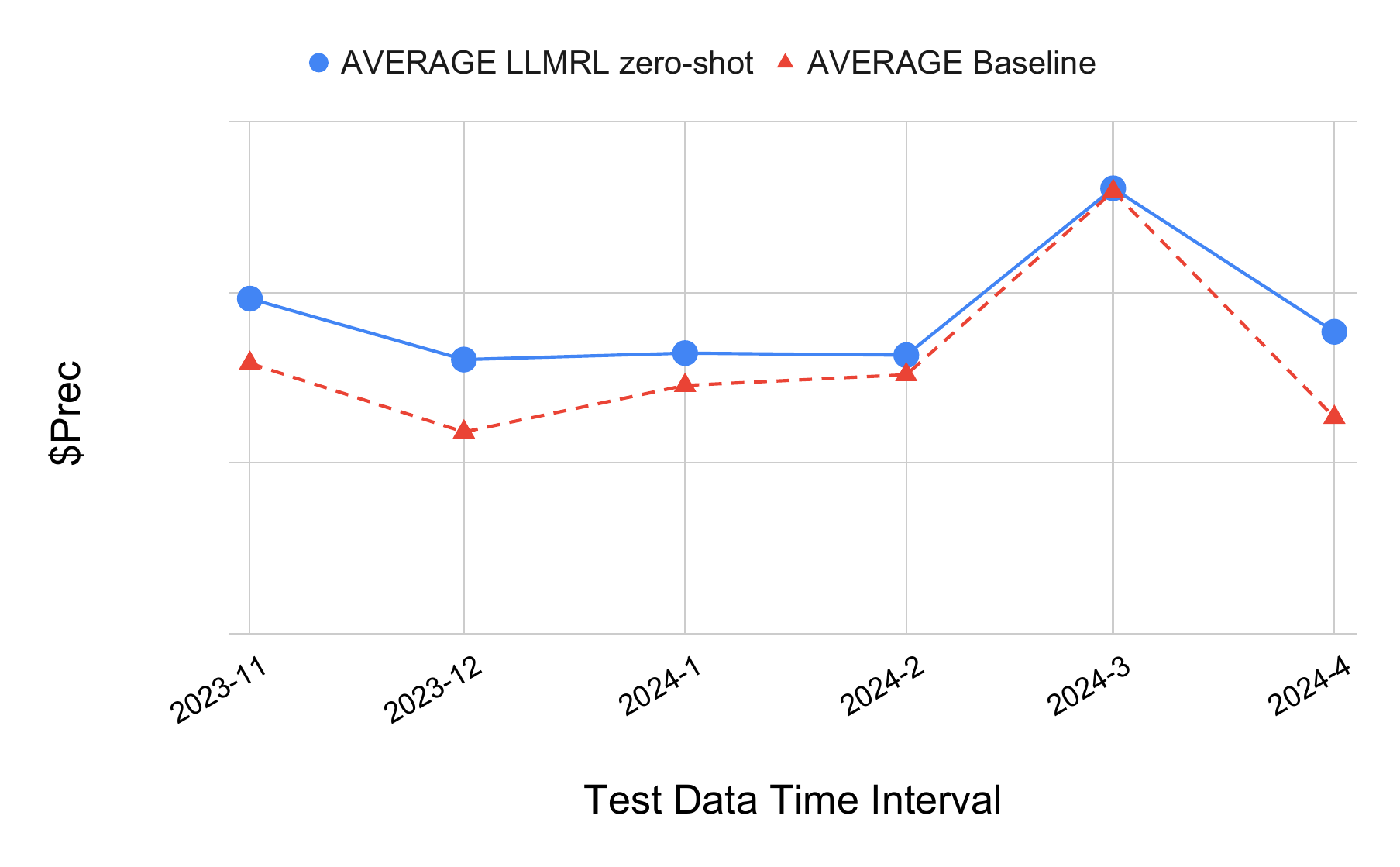}  
        \caption{Averaged blocking \$Prec@\$Recall from 3 LLM guided RL agents, in the zero-shots scenario, on Test L.}
        \label{fig:LLMRL_LT_zs}
    \end{minipage}
\end{figure}

Figure \ref{fig:LLMRL_LT_fs} shows RL agents consistently outperforming the baseline over time. Figure \ref{fig:LLMRL_LT_zs} illustrates zero-shot scenarios where RL agents maintained superior performance.

These evaluations highlight our LLM-enhanced RL framework's durability and effectiveness in real-world applications, supporting continuous deployment without frequent retraining. More results are in Appendix \ref{appendix:long-term eval with different LLM}.

\subsection{Production Efficiency}
Due to the compact architecture and lightweight design of the RL agent network described above, the model supports efficient deployment across both transaction stages. In production, it achieves inference latencies of less than 50 milliseconds using standard CPU infrastructure, making it suitable for real-time fraud detection at scale.

\section{Conclusion}
This study introduces an RL and LLM integration framework for e-Commerce fraud detection, conceptualizing risk assessment as an MDP and enabling dynamic sequential strategies. Our approach, using LLMs to refine reward functions, surpasses traditional human-designed functions in efficiency and zero-shot capability. Empirical tests confirm its superiority over our conventional SL model, with six-month evaluations demonstrating robust performance. The lightweight architecture, is practical for industrial adoption. Future work includes generalizing to more sequential scenarios of risk prevention, and exploring online RL.

\bibliography{custom}

\appendix
\section{Prompts Design for the LLM RL framework}
\label{appendix:Prompts Design}
In this section, we provide the prompts of our LLM RL framework.

\begin{tcolorbox}[title=Prompt 1: Initial Instruction Prompts, colframe=blue, colback=white, breakable]

You are a reward engineer trying to write reward functions to solve reinforcement learning tasks as effective as possible.
Your goal is to: 
(1) write a reward function for the environment that will help the agent learn the task described below. 
(2) try to write improved or try different parameters in the reward function comparing to the reward functions found so far, based on analyzing the provided reward function feedback information below.
The goal of my task is: Design a reward function that enables the RL agent to make more effective decisions across 2 steps for improved overall performance in identifying and blocking risky transactions comparing to a baseline scores in the 1st step, my codes framework of input data as states and train my policy is shown in the code: ```python \{...\} ```.
\end{tcolorbox}

\begin{tcolorbox}[title=Prompt 2: Code Generation Instruction Prompts, colframe=blue, colback=white, breakable]

Your reward function should use useful variables from my codes framework as inputs. As some examples,
here are some examples reward functions proposed by human: ```python \{...\} ```,
and here is the best reward function signature so far: ```python \{...\} ```
Since the reward function will be decorated with @torch.jit.script,
please make sure that the code is compatible with TorchScript (e.g., use torch tensor instead of numpy array). 

Make sure any new tensor or variable you introduce is on the same device as the input tensors. 
The output of the reward function should consist:

    (1) the completed reward function.
    
    (2) the reward code's input attributes must follow the format:"def get\_reward(current\_step,action,target,wgt):".
    
    (3) the code output should be formatted as a python code string: "```python ... ```".
    
    (4) if you have extra functions defined in the reward function, also output these functions completely in one code block.
    
    (5) your codes and the related annotations must be consistent.
    
    (6) it is encouraged to only output your completed reward function python codes in the beginning of your outputs, for the ease of code extraction.
    
    (7) remember to use the backslash properly as a line continuation where you separate one logic line into multiple physical lines for better readability.
    
\end{tcolorbox}

\begin{tcolorbox}[title=Prompt 3: Additional Reward Generation Instruction Prompts with Domain-Specific Context, colframe=blue, colback=white, breakable]

information of the get\_reward:

def get\_reward(current\_step,action,target,wgt):

    \# current\_step is one integer;
    
    \# if the agent is in step 0, then current\_step == 0;
    
    \# if the agent is in step 1, then current\_step == 1;
    
    \# current\_step either equals 0 or 1 in get\_reward function;
    
    \# action and target and wgt are tensors in size (transaction\_batch\_size,);
    
    \# element in action either equals 0 or 1;
    
    \# action == 1 means the transactions that were taken blocking action, action == 0 means the transactions that were taken pass action;
    
    \# element in target either equals 0 or 1;
    
    \# target == 1 means the transactions that are tagged as fraud risk, target == 0 means the transactions are not tagged as risk;
    
    \# wgt is the tensor of dollarwise weight for each transaction.;
    
    \# e.g. ((action==1) \& (target==1) * wgt) means the tensor that have the True Postive GMV value where (action==1) \& (target==1);
    
    \# e.g. ((action==1) \& (target==0) * wgt) means the tensor that have the False Positive GMV value where (action==1) \& (target==0);
    
    \# e.g. ((action==0) \& (target==0) * wgt) means the tensor that have the True Negative GMV value where (action==0) \& (target==0);
    
    \# e.g. ((action==0) \& (target==1) * wgt) means the tensor that have the False Negative GMV value where (action==0) \& (target==1);

    \# the general goal of this reward function is to drive the agent to increase True Postive GMV and True Negative GMV, decrease False Positive GMV and False Negative GMV;
    
    \# this reward function need to drive the agent to block more potential True Postive GMV at the current\_step == 0 than at the current\_step == 1;
    
    \# the returned reward also need to be a tensor in size (transaction\_batch\_size,) or (transaction\_batch\_size,1) , it will be aggregated outside this get\_reward function
    
   return reward
    
\end{tcolorbox}

\begin{tcolorbox}[title=Prompt 4: Feedback Prompts, colframe=blue, colback=white, breakable]

We trained a RL policy using the new found reward function code and tracked my focused metric feedback from a out-of-date test data: 

1. RL Agent Training info: after training in \{...\} episodes, the final blocking action number of the RL agent in first step is: \{...\}, and the final blocking action number of second step is: \{...\}, and the final reward value is: \{...\} comparing to the initial reward value is: \{...\}. Normally we hope to observe the RL agent take more blocking action in the first step than in the second step, and the final reward value should be larger than the initial value.

2. Test evaluation info: after 2 steps actions of a policy agent, we observed the final best precision performance by the agent under some targeting recall thresholds levels: \{...\} and compare with the baseline model, the goal is have better precision compare to the baseline model. The detail of the observations are: Our 2 steps policy agent can reach the similar recall:\{...\} and the agent can reach at best the precision: \{...\}. \
Moreover, the ratio between the bad GMV blocked by first step and the bad GMV blocked by second step is: \{...\}/\{...\}, and the ratio between the total GMV blocked by first step and the total GMV blocked by second step is \{...\}/\{...\};

Error occurred during training: \{...\}

Error occurred during evaluating: \{...\}
\end{tcolorbox}

\begin{tcolorbox}[title=Prompt 5: Reflection Prompts if No Usable Reward Function Found, colframe=blue, colback=white, breakable]

However, after an iteration of reward designs and validations, all of your designed reward functions failed in either training or evaluation, your designs and their regarding failure info are listed here: \{...\}

With all the feedback information, reflect the failed experience regards to your reward functions and output a detailed guidance of reward function design for yourself briefly, in less than length of 1000 tokens:
\end{tcolorbox}

\begin{tcolorbox}[title=Prompt 6: Reflection Prompts if A Better Reward Function Found, colframe=blue, colback=white, breakable]

The previous best reward function's policy agent performance: when the recall threshold is \{...\}, the baseline model can reach the precision: \{...\}. A better new found reward function in iteration \{...\}:\{...\}.
\end{tcolorbox}

\begin{tcolorbox}[title=Prompt 7: Reflection Prompts if A Sub-optimal Reward Function Found, 
colframe=blue, colback=white, breakable]
\small
You found a sub-optimal new reward function in iteration \{...\}:\{...\}, which has worse performance than the previously best reward function.
\end{tcolorbox}

\section{Long-term Test evaluations with different LLM}
\label{appendix:long-term eval with different LLM}
\setcounter{figure}{0} 
\renewcommand{\thefigure}{B.\arabic{figure}} 
In this appendix, we present detailed figures illustrating the performance of different models evaluated in this study.

\begin{figure}[ht]
\centering
\includegraphics[width=\linewidth]{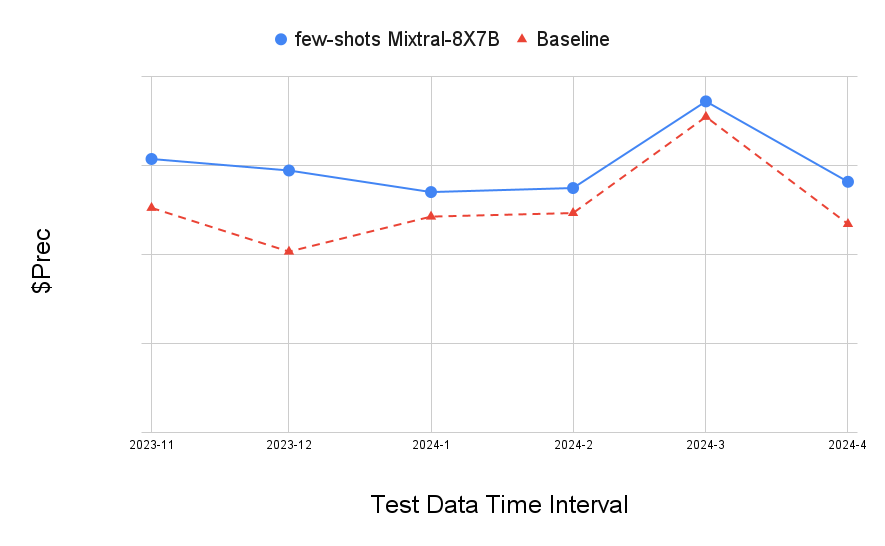}  
\caption{Averaged blocking \$Prec@\$Recall from Mixtral-8X7B guided RL agents, in the few-shots scenario.}
\label{fig:appendix_mixtral_fs}
\end{figure}

\begin{figure}[ht]
\centering
\includegraphics[width=\linewidth]{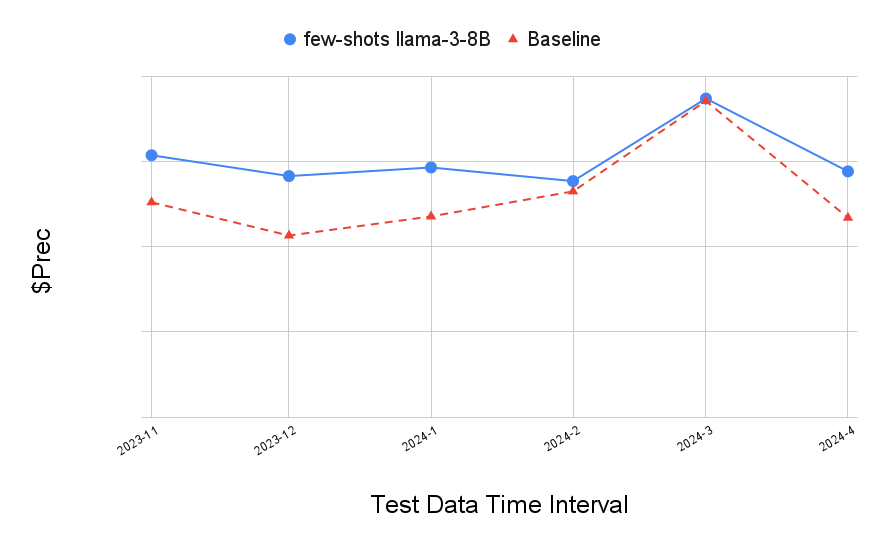}  
\caption{Averaged blocking \$Prec@\$Recall from LLaMa-3-8B guided RL agents, in the few-shots scenario.}
\label{fig:appendix_LLaMa_fs}
\end{figure}

\begin{figure}[ht]
\centering
\includegraphics[width=\linewidth]{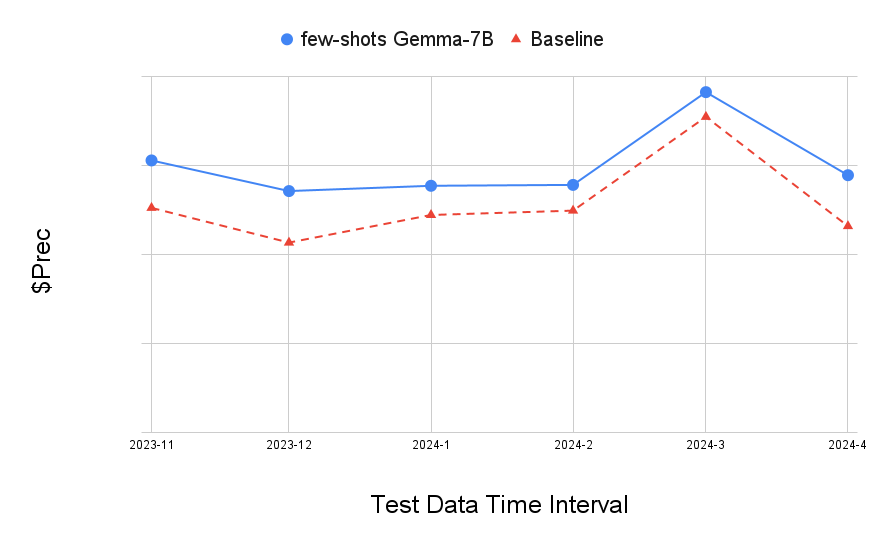}  
\caption{Averaged blocking \$Prec@\$Recall from Gemma7B guided RL agents, in the few-shots scenario.}
\label{fig:appendix_gemma_fs}
\end{figure}

\begin{figure}[ht]
\centering
\includegraphics[width=\linewidth]{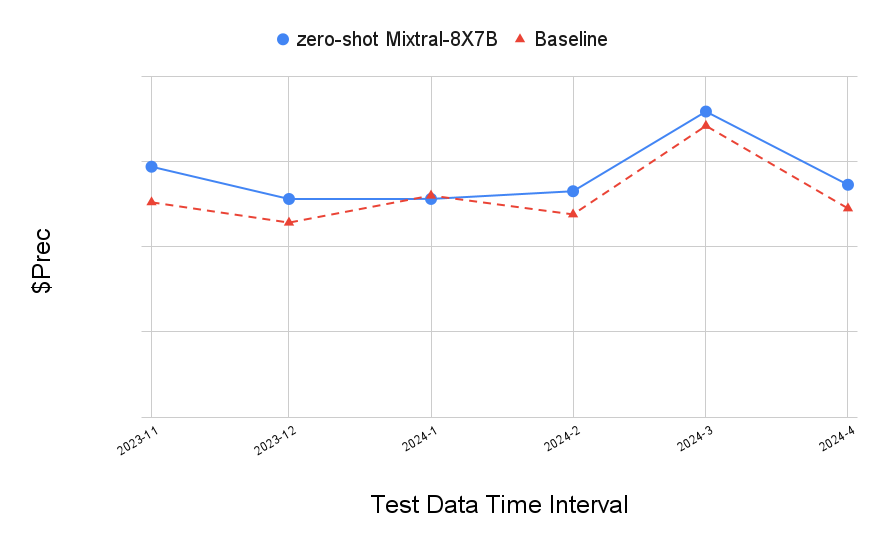}  
\caption{Averaged blocking \$Prec@\$Recall from Mixtral-8X7B guided RL agents, in the zero-shot scenario.}
\label{fig:appendix_mixtral_zs}
\end{figure}

\begin{figure}[ht]
\centering
\includegraphics[width=\linewidth]{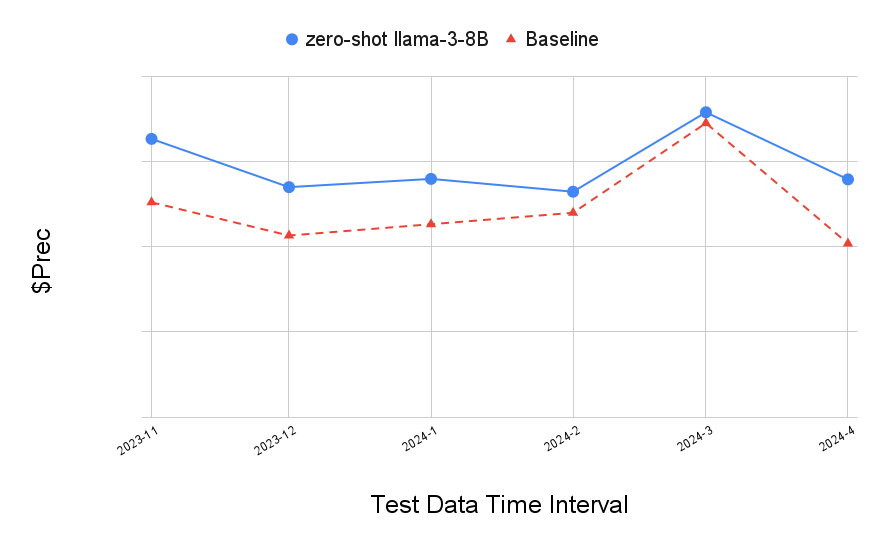}  
\caption{Averaged blocking \$Prec@\$Recall from LLaMa-3-8B guided RL agents, in the zero-shot scenario.}
\label{fig:appendix_LLaMa_zs}
\end{figure}

\begin{figure}[ht]
\centering
\includegraphics[width=\linewidth]{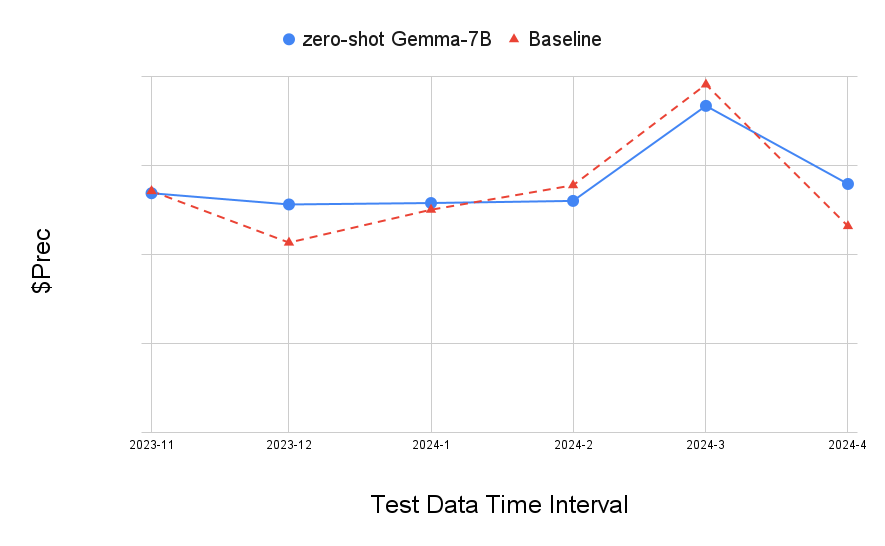}  
\caption{Averaged blocking \$Prec@\$Recall from Gemma7B guided RL agents, in the zero-shot scenario.}
\label{fig:appendix_gemma_zs}
\end{figure}

\end{document}